\pdfoutput=1

\documentclass[11pt]{article}

\usepackage[preprint]{acl}

\usepackage{times}
\usepackage{latexsym}
\usepackage[T1]{fontenc}

\usepackage[utf8]{inputenc}

\usepackage{microtype}
\usepackage{comment}
\usepackage{inconsolata}
\newcommand{\q}[1]{``#1''}
\newcommand{\s}[1]{\textbf{#1}}
\newcommand{\ds}{FEVERFact}

\usepackage{graphicx}
\usepackage{caption}
\usepackage{subcaption}
\usepackage{multirow}
\usepackage{float}
\makeatletter
\newcommand\footnoteref[1]{\protected@xdef\@thefnmark{\ref{#1}}\@footnotemark}
\makeatother

\title{Claim Extraction for Fact-Checking: Data, Models, and Automated Metrics}

\author{Herbert Ullrich \\
AI Center @ CTU FEE\\
Charles Square 13,\\
Prague, Czech Republic\\
\texttt{ullriher@fel.cvut.cz} \\\And
Tomáš Mlynář \\
AI Center @ CTU FEE\\
Charles Square 13,\\
Prague, Czech Republic\\
\texttt{mlynatom@fel.cvut.cz} \\ \\\And
Jan Drchal \\
AI Center @ CTU FEE\\
Charles Square 13,\\
Prague, Czech Republic\\
\texttt{drchajan@fel.cvut.cz} \\}
\begin{document}
\maketitle

\begin{abstract}
  In this paper, we explore\footnote{Full code and data supplement available at \url{https://github.com/aic-factcheck/claim_extraction}} the problem of Claim Extraction using one-to-many text generation methods, comparing LLMs, small summarization models finetuned for the task, and a previous NER-centric baseline QACG.
  As the current publications on Claim Extraction, Fact Extraction, Claim Generation and Check-worthy Claim Detection are quite scattered in their means and terminology, we compile their common objectives, releasing the \ds{} dataset, with 17K atomic factual claims extracted from 4K contextualised Wikipedia sentences, adapted from the original FEVER.
  We compile the known objectives into an Evaluation framework of: \textit{Atomicity, Fluency, Decontextualization, Faithfulness} checked for each generated claim separately, and \textit{Focus} and \textit{Coverage} measured against the full set of predicted claims for a single input.
  For each metric, we implement a scale using a reduction to an already-explored NLP task.
  We validate our metrics against human grading of generic claims, to see that the model ranking on $F_{fact}$, our hardest metric, did not change and the evaluation framework approximates human grading very closely in terms of $F_1$ and RMSE.
\end{abstract}

\section{Introduction}\label{sec:intro}

Recent research in Natural Language Processing has extensively covered the case of automated fact-checking as a pipeline of retrieval and inference tasks~\cite{fever-2022-fact,thorne-vlachos-2021-evidence}.
As noted in other research~\cite{guo-etal-2022-survey}, these steps alone do not represent the whole challenge human fact-checkers face, notably omitting the step of coming up with the claims to be checked in the first place.
\begin{figure}[h]
    \centering
    \includegraphics[width=\columnwidth]{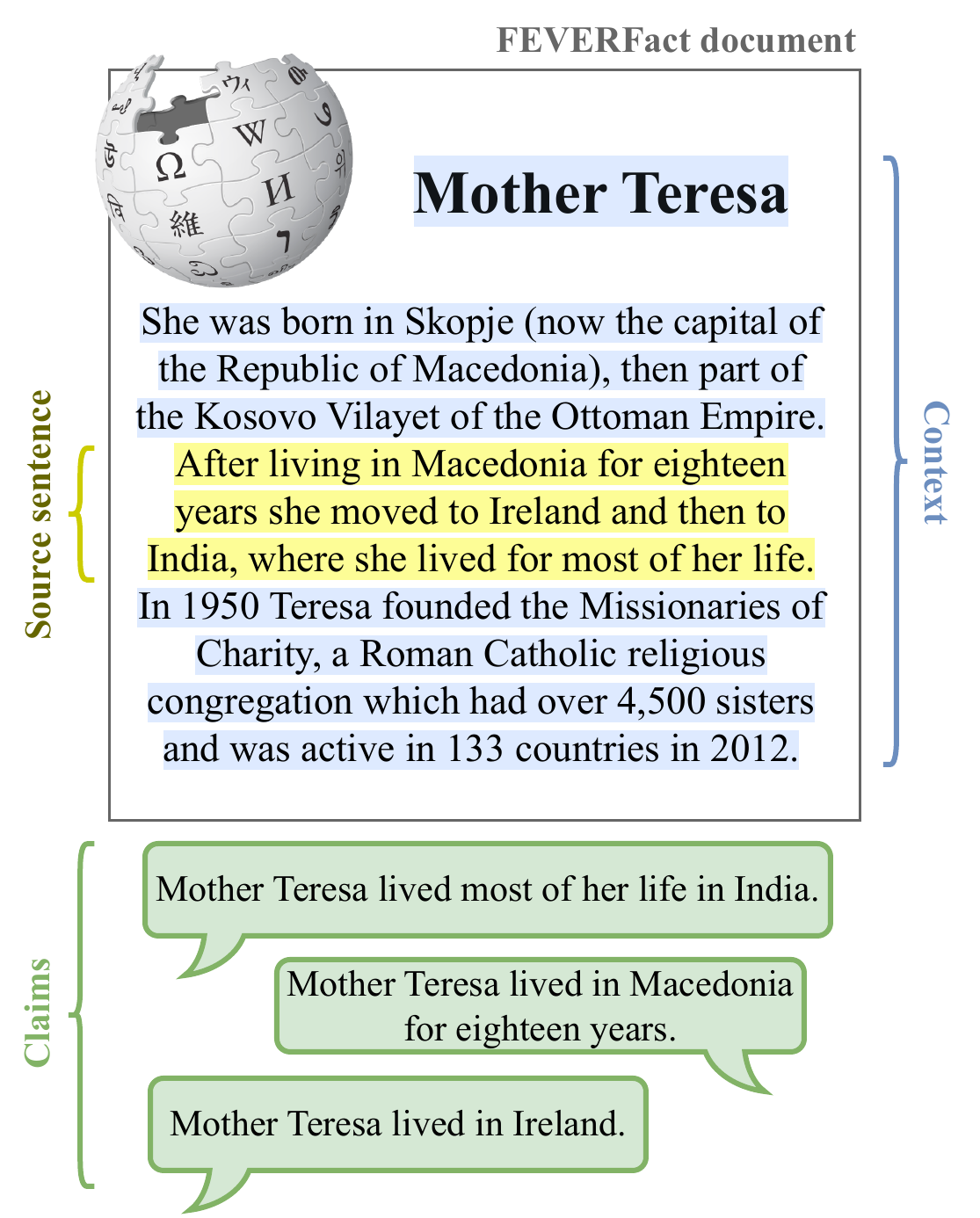}
    \caption[An example from \ds{} dataset]{An example of claim extraction from the \ds{} dataset constructed from data published by~\citealt{thorne-vlachos-2021-evidence}}
    \label{fig:feverfact}
\end{figure}

The real-world factual claims may be scattered throughout a politician's social network post, pieced together from a debate transcript, distilled from a lengthy news article, etc. 
Research such as~\cite{barroncedeno2020overview} therefore proposes augmenting the pipeline by the step of \textit{claim detection}, commonly modelled as classification of \textit{check-worthy} claims within a real-world text segmented into its sentences.

The claim detection paradigm, however, only kicks the can down the road -- the detected \q{check-worthy} parts of text still need to be processed by a human fact-checker~\cite{sheikhi-etal-2023-automated} and re-written into the form of a \textit{claim}: a factual statement presenting the source text information without the need of additional context in a single sentence.

The automated generation of such claims is desirable not only for the fact-checkers themselves~\cite{deng-etal-2024-document}, but also for further research in the field~\cite{drchal2023pipeline} where the generation of claims on top of large-scale corpora was shown to yield valuable data to train models~\cite{pan-etal-2021-zero}.

In this paper, we explore the abstractive methods of claim extraction, modelled as one-to-many text-to-text task: after seeing the whole input text, the model is tasked to formulate a set of claims it makes, as simple self-contained sentences faithful\footnote{An interesting close parallel for this task are the single-sentence summarization methods (also known as the \textit{extreme summarization}) explored before in~\citealt{narayan-etal-2018-dont,mao-etal-2022-citesum}, finetuning of which we examine as one of the approaches in Section~\ref{sec:methods}.} to the source. 
We construct a downstream task for the models -- the \textbf{\ds} dataset -- and compile the criteria previously studied for extracting factual claims~\cite{wright-etal-2022-generating,10.1007/978-3-031-28241-6_59}, proposing methods of their automatic evaluation at a scale.

This paper explores the claim extraction using generative methods in breadth, accumulating a set of data, automated reference-based and reference-free metrics and proposing solvers for the task, yielding a complete scheme of the automated generation of check-worthy claims and measuring their qualitative properties to identify possible problems of each claim.

\subsection{Main Contributions}
\begin{enumerate}
    \item We publish the \ds{} dataset (Figure~\ref{fig:feverfact}) with 4.4K contextualized Wikipedia sentences, and a total of 17K check-worthy claims extracted from them by annotators.
    \item We propose an automated evaluation framework for claim extraction reusable for other generative fact-critical tasks, consisting of 6 metrics: Atomicity, Fluency, Decontextualization, Faithfulness, Focus and Coverage.
    We compile current research in the field to name the relevant metrics, as well as to find their scoring methods, combined with novel ideas where appropriate (see Section~\ref{sec:metrics}).
    \item We explore generative methods for extracting the \ds{} claims -- QACG from~\cite{pan-etal-2021-zero}, LLM prompting, and LM transfer learning, publishing our tuned models.  
    \item We annotate a blinded sample of generated claims to validate our evaluation framework and challenge the benchmarks it produces.
\end{enumerate}

\section{\ds{} Dataset}\label{sec:feverfact}
In order to obtain a dedicated dataset, we utilize the \textbf{FEVER Error Correction} data published in~\cite{thorne-vlachos-2021-evidence}.
Despite the dataset being designed for a different task, it releases a set of 17.5K truthful claims \textit{directly extracted} from a sample of sentences of the 2017 Wikipedia corpus.
The annotation of these claims dates back to the WF1a task of the original \textbf{FEVER}~\cite{thorne-etal-2018-fever}, in which the annotators were instructed to extract 2--5 claims from a source sentence, being provided with its neighbouring sentences and page title for context, as shown in Figure~\ref{fig:feverfact} -- hereinafter, we refer to this whole body of text as to the \textit{\ds{} document}.
Years later, the original source sentence each claim was extracted from, however, may not be reproduced from the data, as each sentence is only represented by an integer ID pointing to a mapping file within a deleted Amazon bucket, lost to its authors\footnote{As per direct Slack communication with FEVER authors.}. 

With a simple idea, we manage to reverse-engineer this mapping:
we first group the claims by the ID of their (unknown) source sentence, taking the most common Wikipedia article among each group's FEVER gold evidence.
This should be the article group was extracted from, since all the WF1a claims were annotated to be true and only contain the information from their source sentence, without any additional knowledge.
We then, sentence-by-sentence, scan this article using a \textit{Natural Language Inference (NLI)} model, looking for a sentence which entails the most of the claims in the group, tossing the groups where no claim is entailed by any sentence due to noise in our method.
Doing so using a \texttt{nli-deberta-v3-small}~\cite{he2021deberta,he2021debertav3,reimers-gurevych-2019-sentence} pre-trained CrossEncoder yields near-ideal results -- our annotation on a sample of 2\% (89 sentences, 350 claims) datapoints shows that in 94.4\% cases the \textit{entire} group of claims could be directly extracted from the reverse-engineered source sentence, only using its neighbouring sentences and page title to resolve coreferences.

Thus we release dataset we call \textbf{\ds{}}: 4.4K 3-sentence documents (plus a page title), each annotated with a set of claims that may be extracted from its middle sentence for a total of 17K atomic factual claims.
We split it into train, dev and test set in an 80:10:10 ratio, preventing the same page title from appearing in two different splits.

\subsection{\ds{} recall}
In the wild, an ideal dataset for claim extraction would feature \textit{all} check-worthy claims that can be extracted from given source sentence or even from the whole \ds{} document.
Such data is extremely hard to annotate due to the requirement for claim atomicity\footnote{Each claim describing a single entity or relation, see Section~\ref{atomicity}} -- even relatively simple grammatical paralellisms such as zeugmata or compound sentences can explode the number of relations that can be extracted from the sentence.

Even so, having a large-enough sample of claims matched to a source text would yield a good approximation on which information within the text is check-worthy enough to be extracted.

The original annotators behind data used in \ds{} were tasked to extract 2--5 check-worthy claims from the source sentence, producing a median of 4 claims per source.
To probe whether this number is enough, we perform a Named Entity Recognition (NER) on both \ds{} claims and their source sentences using NameTag~3~\cite{strakova-etal-2019-neural}, to find that the recall of named entity words in FEVER claims is about 67\% (taking source entities as a reference).
While such metric is on the rough side and future annotation is desirable, we conclude that the \ds{} claims cover most of the information and are suitable enough to be used as a reference for reference-based claim extraction metrics (Section~\ref{reference-based}).

\section{Claim Extraction Models}\label{sec:methods}
Using the \ds{} dataset to train and evaluate claim generators, we model the automated claim extraction as a \q{one-to-many} sequence-to-sequence task~\cite{sutskever-2014} -- given a single \ds{} document, predict a set of facts it claims.
We experiment with prompting, transfer learning and a NER-based baseline used on FEVER data previously in literature.

To all the models, we do not provide any distinction (such as separators) between the \ds{} source sentence and its context, feeding the whole \ds{} document (Figure~\ref{fig:feverfact}) to each model's input.
This is done experimentally, to give idea whether different models are able to learn to only focus on the information relevant to the user (in our case, arbitrarily, only the information claimed by the middle sentence) just from its training/few-shot examples.
If so, this primitive example of \q{positional} relevance of input information could motivate collection of claim extraction datasets capturing more challenging notions of relevant information to extract, such as check-worthiness of claims scattered through a political debate or Twitter discourse, where, unlike in encyclopedic style, only a small proportion of text is interesting for checks~\cite{10.1007/978-3-031-28241-6_59}. 

We experiment with the following models:

\begin{enumerate}
    \item \textbf{Question Answering for Claim Generation (QACG)}~\cite{pan-etal-2021-zero} is an off-the-shelf baseline we use.
    It has been built dedicatedly to generate claims: given an input text, its named entities are extracted, fixing each entity $e$ as an answer for a question $Q$ to be generated.
    The $(Q,e)$ is then converted into a declarative sentence using a QA-to-claim model, arriving to one factual claim for each of the source-text entities.

    While being a valuable baseline, the entity-centric approach has its caveats: the QACG pipeline has many steps, each a language model and, therefore, a point of failure -- mistake in any step propagates to the result.
    The QACG also lacks trainable parameters for \textit{choosing the appropriate} claims.
    If, for example, one wants to generate claims from a debate, it does not have mechanisms to learn to omit recurring irrelevant, entity-dense guest introductions. 
    \item \textbf{LLMs and few-shot learning}~\cite{NEURIPS2020_1457c0d6} has been recently a popular, universally well-performing solution to be reckoned with.
    Models such as GPT-4~\cite{openai2023gpt4} or Mistral-instruct~\cite{jiang2023mistral} have been shown to adapt well for similar tasks.

    We examine \texttt{gpt-4-turbo} in a 3-shot setting and finetune \texttt{Mistral-7b-instruct-v0.2} model on our \ds{} train data using the quantized QLoRA~\cite{dettmers2023qlora} approach with $r=64, \alpha=32$ on 4 bits.

    The GPT-4, however, is a blackbox, and the open-source LLMs are computationally expensive.
    This motivates an examinaton of other methods to be reproducible on much lower resources using easy-to-obtain data.
    \item \textbf{T5 transfer learning}~\cite{2020t5}: our small model of choice (based on the preliminary experiments) is the \texttt{t5-small-finetuned-xsum}\footnote{Initial weights loaded from openly available \url{https://huggingface.co/darshkk/t5-small-finetuned-xsum}}
    
    The T5 was pre-trained on various text-to-text tasks using a \textit{span-corruption} objective, making it adapt well to our task despite a relatively small (3.5K \ds{} documents, 13.5K claims) train size.
    We have found that fine-tuning it on a single-sentence summarization dataset such as XSum~\cite{narayan-etal-2018-dont} \textit{and then} on \ds{} yields even better results, possibly due to similar task definition.

    Two approaches were examined: first, where T5 was trained to output a concatenation of all claims in a single prediction\footnote{The claims were then separated using PySBD~\cite{sadvilkar-neumann-2020-pysbd}}, second, where the T5 was tuned to output a \textit{single} claim as a sentence (using our data as 13.5K document-claim pairs).
    In the prediction stage, the latter approach was coupled with \textit{diverse beam search}~\cite{vijaykumar_beam} decoding technique to generate arbitrary number ($k$) of single-sentence claims per text, using $k$ beam groups and a diversity penalty of $1$.
    Its results on our task being competitive with the other models are particularly encouraging, as the single-sentence summarization data is easily available in other settings and languages (just XLSum~\cite{hasan-etal-2021-xl} features data in 45 languages expertly annotated for the task). 
\end{enumerate}
\section{Evaluation Framework: Claim Metrics}\label{sec:metrics}
To evaluate our models, we compile claim quality criteria introduced in other research.
Each predicted claim is checked to be \textit{faithful}, \textit{fluent}, \textit{atomic} and \textit{decontextualized}.
A set of claims extracted from the same document is measured to \textit{focus} solely on its relevant information and \textit{cover} all of it.
While the criteria above can be checked by human grading as in~\cite{wright-etal-2022-generating}, research such as~\cite{wang-etal-2020-asking,ffci} already proposes ways of how to evaluate some of the qualities automatically, alleviating the need of human grading of model outputs.
In this section, we attempt to compile known evaluators for our task, reducing each metric to an already well-explored NLP challenge, proceeding to validate our metrics against real human annotations on generated claims.

\subsection{Reference-free evaluation metrics}
\label{atomicity}
For the reference-free metrics which score each claim on its own, we use the criteria named by~\citealt{wright-etal-2022-generating}: Atomicity, Fluency, Decontextualization, and Faithfulness, adapted from \citealt{10.1007/978-3-642-38288-8_33}'s AIDA (Atomic, Declarative, Independent, Absolute).
To allow automated evaluation at a scale, we find following reductions to known NLP tasks to replace human grading:

\begin{enumerate}
    \item \textbf{Atomicity} -- \textit{does the claim describe a single entity, relation, or process?}

    While the real-world factual claims are not often atomic in the true sense (i.e., the claims are typically more convoluted than \q{A is B} or \q{C does D}), breaking a more complex factual statement into a set of atomic claims trivializes the inference on top of such claims, as well as allows certain level of explainability~\cite{dammu-etal-2024-claimver}, such as which parts of complex statement disinterpret which facts.
    
    We propose the following scheme of atomicity classification of claim $c$, reducing it to a Relationship Extraction task as:

    $$A(c) := \left\{ 
        \begin{array}{ c l }
            1 & \quad \textrm{if } |RE(c)| \leq 1 \\
            0                 & \quad \textrm{otherwise}
        \end{array}
    \right.$$

    Where the relation-extraction result is interpreted as a set of undirected relations $RE(c) = \{\{s_1,t_1\}\dots\}$, in order to avoid counting symmetrical relationships like $(Trump, president\_of, USA)$, $(USA, governed\_by, Trump)$ twice.

    For the $RE$ solver, we recommend using $\textrm{REBEL}$~\cite{huguet-cabot-navigli-2021-rebel-relation} due to its manageable size and end-to-end approach -- other models may rely on entity-pair extraction and classification~\cite{yamada-etal-2020-luke} -- we proceed to use its Large variant.

    We also experimented with a non-binary metric $A'(c)=\frac{1}{max\{1,|RE(c)|\}}$, but during our experiments, non-atomic model outputs were quite rare, so punishing them by zeroing-out their $A$ score worked best for averaging across large number of claims.

    \item \textbf{Fluency} -- \textit{is the claim grammatically correct and intelligible?}
    
    The task of fluency, also referred to as \textit{grammaticallity}, is well studied in literature, with most recent research modelling it as a grammatical error detection (GED) and correction (GEC) tasks.
    
    The available techniques range from simpler ridge regression models based on linguistic features~\cite{heilman-etal-2014-predicting} through using syntactic log-odds ratio (SLOR)~\cite{kann-etal-2018-sentence} or perturbing the claims words and characters to find local optima in the output probability using a language model such as GPT-2 as its reference~\cite{yasunaga-etal-2021-lm} to promping a LLM (such as GPT-3) to obtain a model-inferred score using few- or zero-shot learning~\cite{fu-etal-2024-gptscore}. 
    
    The best performing approach we studied revolves around CoEdIT~\cite{raheja-etal-2023-coedit} GEC model, coupled with Scribendi score~\cite{islam-magnani-2021-end} to rate the improvement between each claim and its CoEdIT correction.
    Scribendi score combines perplexity scores and Levenshtein distance, yielding -1 for bad correction, 0 for no improvement, and +1 for correction improving the claim.
    While this metric is originally designed for the evaluation of GEC systems, it performs well on rough fluency rating, looking for zero or negative improvement in CoEdIT corrections:
    $$G(c):=\left\{ 
        \begin{array}{ c l }
        1 & \textrm{if } \textrm{\footnotesize{Scribendi}}(c,\textrm{\footnotesize{CoEdIT}}(c)) \leq 0 \\
        0 & \textrm{otherwise}
    \end{array}
    \right.$$

    \item \textbf{Decontextualization} -- \textit{can the claim be correctly interpreted without any additional context from the source document or elsewhere?}
    \citealt{choi-etal-2021-decontextualization} proposes the decontextualization as a text-to-text task, training a T5 to receive context and a sentence on its input, outputting the decontextualizated sentence, resolving its pronouns into proper nouns and relative terms into absolute according to the given context marked with separators.

    As with atomicity, a strict binary classification worked best, as the non-decontextualized model outputs were rare.
    For context, we use the full \ds{} document $d$:
    $$D(d,c) := \left\{ 
        \begin{array}{ l l }
            1 &  \textrm{if } \textrm{ T5}_{\textrm{\tiny{d}}}(d,c) = c \\
            0                 &  \textrm{otherwise}
        \end{array}
    \right.$$
    where $\textrm{T5}_{d}(d,c)$ denotes the output of $\textrm{T5}_{\textrm{\tiny{large}}}$ model trained by~\citealt{choi-etal-2021-decontextualization}.

    \item \textbf{Faithfulness} -- \textit{does the claim contain only information consistent with its source?}
    
    Faithfulness has been extensively studied on its own to detect hallucination in text-to-text tasks. \textsc{AlignScore} claimed a state of the art in~\citealt{zha-etal-2023-alignscore}, looking for optimum alignment of output and input parts, in terms of a RoBERTa~\cite{liu2019roberta} classifier with a \textit{unified alignment function}.
    While compact (125M parameters in \textit{base} version), it outperforms metrics based on GPT-4 that is orders of magnitude larger~\cite{zha-etal-2023-alignscore}.
    $$Faith(d,c) := \textsc{AlignScore}_{base}(d,c)$$

\end{enumerate}

\subsection{Reference-based evaluation metrics}\label{reference-based}
The evaluation metrics introduced above do not rely on any gold data for their reference.
But what if we need to evaluate the whole group of claims, to see if it captured all the information it should and nothing more?

The notion of what a good set of claims is varies task-to-task, so engineering rules may do more harm than good.
Let us therefore use examples, like~\citealt{ffci} suggests in their FFCI framework for interpretable summarization metrics, measuring \textit{Focus} and \textit{Coverage} to see whether the model extracts what a human would.
Besides the predicted set of claims $C$, which serves as the input for multi-claim metrics, assume a set of gold claims $G$ extracted from the same document -- in \ds{}, those would be the full sets of claims (green in Figure~\ref{fig:feverfact}) obtained from annotation.

\begin{enumerate}
    \setcounter{enumi}{4}
    \item \textbf{Focus} -- \textit{what is the proportion of gold (relevant) information among all the information listed in the generated claims?}
    
    The metric is analogous to the concept of \textit{precision}: $\frac{|true~positives|}{|positives|}$, but rather then exact matching, we seek a measure of semantic overlap.
    FFCI uses QAGS~\cite{wang-etal-2020-asking}, which uses a Question Generation model (QG) to formulate questions in natural language based on all $|C|$ predicted claims. The questions are then twice answered using a Question Answering (QA) model, giving it knowledge from (i.) the predicted claims (ii.) the gold claims written by a human.
    Focus is then defined as the proportion of questions with the same answers extracted from the gold and predicted claims among all questions the model can generate from the predicted claims.
    In our experiments, this metric was too noisy, with QG, QA and aggregation faults all propagating into the final result.

    To propose a simpler method, we exploit that the gold claims in $G$ can be concatenated into a mock-document.
    Since gold claims are decontextualized by annotation, this preserves their meaning, reducing Focus into $|C|$ independent tasks of deciding Faithfulness:
    $$Foc(G,C):=\frac{1}{|C|}\sum_{c\in C}Faith(concat(G), c)$$
    We also encourage experiments with \textit{claimwise focus}, computing the single probability $Foc(G,\{c\})$ for each pred. claim $c$ separately, to see \textit{which exact} claim in $C$ should be extracted according to $G$ and which should not.
    \item \textbf{Coverage} -- \textit{what proportion of gold (relevant) information from the source is featured in $C$?}
    
    Can be simply adapted from focus: 
    $$Cov(G,C) := Foc(C,G)$$

    \citealt{ffci} proposes this trick, and while our underlying $Foc$ method differs, the argument swap still works to yield proportion of gold information extracted into $C$. 
    Whether the claims in $C$ are non-decontextualized by a faulty prediction and could influence each other's meanings upon concatenation can be checked for using the metric from Section~\ref{atomicity}, marking the claims to toss, or leave them in if the level of noise is tolerable.
    We went with the latter as $>94\%$ claims were decontextualized upon human grading (Table~\ref{gold-metrics}).
    
    \item \textbf{Redundancy}\footnote{\label{fn:redundancy}Unlike Focus and Coverage, we did not validate this Redundancy metric with annotations, but as its validity stems from the same principles as that of $Foc$ and $Cov$, we feature it anyways for added insights. The metric was motivated by feedback from the annotators of data used for Table~\ref{gold-metrics}} -- \textit{What is the proportion of duplicate information among generated claims?}
    
    the definitions of $Foc$ and $Cov$ inspire an ellegant estimate of \textit{model redundancy} -- if $|C|$ claims are generated by a one-to-many seq2seq model, what proportion of them can be expressed from the \textit{other} model outputs:
    $$Red(C) := \frac{1}{|C|}\sum_{c\in C}Faith(concat(C\setminus c), c)$$ 
\end{enumerate}

\subsection{\textbf{$F_{fact}$-value}}
\label{ffact}
As our experiments in Section~\ref{sec:results} suggest, the $Foc$ and $Cov$ do indeed behave like Precision and Recall metrics in the sense of their mutual tradeoff -- if a na\"ive model scores very high in one (such as QACG covering \textit{every part} of the \ds{} document), a significant decrease can be observed in the other.
    
In our experience, the reference-free single-claim metrics (Section~\ref{atomicity}) did not pose a tough challenge to modern NLP methods, most of their corresponding mean scores in both Table~\ref{auto-metrics} and~\ref{gold-metrics} being very high, raising a concern they may be a too-easy benchmark nowadays.
Provided with the gold annotations for each source document, we therefore suggest that the most important aggregate metric to compare the future claim extraction SotA models should be the \textbf{$F_{fact}$ score} -- harmonic mean of $Foc$ and $Cov$.
\begin{table*}[h]
    \centering
    \begin{tabular}{@{}l|cccc|cccc@{}}
        Model & \footnotesize{Atomicity}   & \footnotesize{Fluency}    & \footnotesize{Decontext.}    & \footnotesize{Faith.} & \footnotesize{Focus} & \footnotesize{Coverage}  & $F_{fact}$& \tiny{Redundancy\footnoteref{fn:redundancy}}  \\
        \hline
        \footnotesize{\texttt{qacg}}                      & 0.89   & 0.69   & 0.70   & 0.88      & 0.20  & 0.67        & 0.30     & 0.44    \\
        \footnotesize{\texttt{gpt-4-turbo-3-shot}}        & 0.92   & 0.70   & 0.77   & \s{0.99} & 0.21  & \s{0.81}     & 0.34     & \s{0.14}\\
        \footnotesize{\texttt{qlora-mistral-instruct}}    & 0.95   & 0.75   & 0.80   & 0.95      & \s{0.58}  & 0.63    & \s{0.61} & 0.19    \\
        \footnotesize{\texttt{t5\_sm\_diverse\_7\_beams}} & 0.95   & 0.74   & 0.80   & 0.91      & 0.55  & 0.58        & 0.56     & 0.59    \\
        \footnotesize{\texttt{t5\_sm\_multi-claim}}       & \s{0.96} & \s{0.76} & \s{0.82} & 0.95  & \s{0.58}  &0.51   & 0.55     & 0.54    \\
        \hline
        \hline
        \multirow{2}{*}{\footnotesize{\textit{Evaluation method}}}      & {\small{\textit{REBEL}}} & \small{\textit{CoEdIT}} & \footnotesize{\textit{T5}} & \footnotesize{\textit{Align-}} & \footnotesize{\textit{Concat}} & \footnotesize{\textit{Concat}} & \footnotesize& \\
        \noalign{\vskip -5pt}
        &   \tiny{$|rel|\leq1$}    & \tiny{\textit{+Scribendi}} &\tiny{\textit{Decontext}} & \footnotesize{\textit{Score}}        & \footnotesize{\textit{AlignS.}} &  \footnotesize{\textit{AlignS.}}  & &   \\
        \footnotesize{\textit{Validation against human}}      & \textit{0.96} & \textit{0.80} & \textit{0.86} & \textit{0.92}  & \textit{0.23} & \textit{0.22}  && \\
        \footnotesize{\textit{Validation method}}      &\multicolumn{4}{c|}{\footnotesize{\textit{$F_1$ --  higher is better}}}   &\multicolumn{4}{c}{\footnotesize{\textit{Root mean squared error --  lower is better}}} \\ 

    \end{tabular}
    \caption{\label{auto-metrics}
        \textbf{Automated claim metric} averages across model-generated claims on \ds{} test set. The best value for each metric marked bold. Model choices and training procedures described in Section~\ref{sec:methods}, claim quality metrics in Section~\ref{sec:metrics}. Automated scoring methods and their validation against human grading (Table~\ref{gold-metrics}) in italic.}
\end{table*}

\begin{table*}
    \centering
    \begin{tabular}{@{}l|cccc|cccc@{}}
        Model & \footnotesize{Atomicity}   & \footnotesize{Fluency}    & \footnotesize{Decontext.}    & \footnotesize{Faith.} & \footnotesize{Focus} & \footnotesize{Coverage} & $F_{fact}$& ~~~~~~~~~~~~~~~~   \\
        \hline
        \footnotesize{\texttt{qacg}}                      & 0.99   & 0.85   & 0.91   & 0.76   & 0.19     & 0.60             & 0.28     &\\
        \footnotesize{\texttt{gpt-4-turbo-3-shot}}        & 0.98   & \s{0.97} & \s{0.96} & \s{0.93} & 0.24     & \s{0.79}   & 0.37     &\\
        \footnotesize{\texttt{qlora-mistral-instruct}}    & \s{1.00} & 0.96   & \s{0.96} & 0.90   & \s{0.60} & 0.69         & \s{0.64} &\\
        \footnotesize{\texttt{t5\_sm\_diverse\_7\_beams}} & 0.99   & 0.89   & 0.95   & 0.79   & 0.51     & 0.62             & 0.56     &\\
        \footnotesize{\texttt{t5\_sm\_multi-claim}}       & \s{1.00} & 0.91   & 0.94   & 0.88   & 0.47     & 0.50           & 0.49     &\\
        \hline
        \hline
        \multicolumn{9}{c}{\footnotesize{\textit{Inter-annotator agreement (sample of $\sim23\%$ annotated claims, $\leq$ 5 annotators per claim, at least 2)}}}\\
        \footnotesize{\textit{Krippendorff's $\alpha$}} & \textit{0.27} & \textit{0.41} & \textit{-0.01} & \textit{0.53} & \textit{0.75} & \textit{0.64} && \\
        \footnotesize{\textit{GWET's AC1}}              & \textit{0.95} & \textit{0.86} & \textit{0.85}  & \textit{0.83} & \textit{0.80} & \textit{0.66} && \\
        \footnotesize{\textit{\%-agreement}}            & \textit{0.87} & \textit{0.86} & \textit{0.97}  & \textit{0.84} & \textit{0.88} & \textit{0.82} && \\

    \end{tabular}
    \caption{\label{gold-metrics}
        \textbf{Blinded human annotation} averages across a sample of 1110 claims generated from 40 (9\%) \ds{} test documents. Annotation of reference-free metrics (left) was done using grading scales (Appendix~\ref{sec:annotation}) adapted from~\citealt{wright-etal-2022-generating} binarizing the best grade to 1, others to 0. Reference-based metrics (right) annotated using a checkbox interface over gold and pred. claims (Appendix~\ref{sec:annotation}). The best value for each metric marked bold. Inter-annotator agreement experiment results in italic.
        Krippendorff's $\alpha$ left in for completeness, albeit very inappropriate for tasks with such class imbalance, as also noted by~\citealt{wright-etal-2022-generating}, Gwet's AC1 being the more appropriate agreement metric for our task.}
\end{table*}

\section{Results}\label{sec:results}

After training the models described in Section~\ref{sec:methods} on \ds{} dataset, we had each model extract a set of claims from each of 444 \ds{}-test documents.
We evaluated the claims using the automated metrics described in previous Section~\ref{sec:metrics}.
To validate the metrics, we have also annotated a subset of 9\% (40) \ds{}-test documents (and 1110 claims extracted from them by the models in total), using a custom annotation platform described in Appendix~\ref{sec:annotation}.

\subsection{Model comparison}
In tables~\ref{auto-metrics} and~\ref{gold-metrics} we use the automated and human-annotated metrics to benchmark our models, on full \ds{}-test set and a sample of 40 documents, respectively.
The benchmarks reveal that all models score high in Atomicity, Fluency, Decontextualization, and Faithfulness reference-free metrics -- confirmed by the human annotations used to compute the Table~\ref{gold-metrics}, these do not appear to be the challenges modern NLP claim-extractors would struggle with.

The reference-based metrics of Focus and Coverage are where significant differences can be found and tradeoffs can be seen -- models with highest coverage focus too little and vice versa. 
Taking their harmonic mean, $F_{fact}$, the fine-tuned models \texttt{qlora-mistral-instruct} and \texttt{t5\_sm} take a significant lead, showing that the models were able to learn that the \ds{} claims come from the middle sentence on input, which is the information annotators were tasked to focus on in the original~\citealt{thorne-etal-2018-fever}'s annotation experiment -- hopefully, this could mean the models could be tuned for trickier notions of check-worthiness, such as those studied by~\citealt{10.1007/978-3-031-28241-6_59}, in future works.

The T5 coupled with \textit{diverse beam search} decoding strategy described in Section~\ref{sec:methods} shows promising results, despite it being essentially an abstractive summarization model.
Its redundancy is the highest (diverse outputs only assured through a decoding strategy), but not a clear outlier.
We find this very encouraging for further use of the diverse beam search predictioning in settings where only one claim per source document is annotated, such as in AVeriTeC-DCE~\cite{deng-etal-2024-document}, to produce multiple interesting results.
Datasets like XLSum~\cite{hasan-etal-2021-xl} with an objective highly similar to claim extraction and good availability of summarization models across languages make it an accessible choice with competitive performance.
While the GPT-4 could have certainly been prompted to fit to the \ds{} claim extraction better than in our study, we still recommend to use the provided \texttt{t5}'s or \texttt{Mistral} as the baseline for future investigations for their size, ease of use, and trainability for whichever claims one wants to extract through examples.   

\subsection{Metric validation}
To validate our evaluation framework, we have annotated the claim properties on grading scales with instructions adapted from~\cite{wright-etal-2022-generating} and the Focus and Coverage using a simple checkbox interfaces shown in attachment~\ref{sec:annotation}.

The human annotations confirm that the reference-free methods (Section~\ref{atomicity}) are easy for today's models.
They, surprisingly, also preserve the leaderboard of models based on $F_{fact}$, the hardest metric, requiring an understanding of which claims are relevant in \ds{} data and punishing greedy approaches.
We evaluate the reference-free metric estimation methods using the $F_1$ score, chosen due to the imbalance of classes on real generated claims (this goes hand in hand with metrics being too easy for models).
For faithfulness and fluency, we binarize the human-annotated grades (the highest grade to 1, others to 0), to be directly comparable with automated metrics (Section~\ref{sec:metrics}) -- AlignScore used for Faithfulness was simply rounded to 0 or 1 (threshold 0.5) for this experiment.
All metrics score above $80\%$ $F_1$, testifying to their soundness and usability at a scale in scenarios where a good approximation is enough.

We proceed to validate our Focus and Coverage metrics against human annotation, using the root mean squared error to obtain values around 0.22, which is also encouraging for further use.
To see whether this value is not accidental (correct proportion, but erroneously distinguished gold claims), we also measure the \textit{claimwise} $Foc$ and $Cov$ to see the per-claim contribution to metrics as a probability $Foc(G,\{c\})$, $Cov(\{g\},C)$,  as shown in Section~\ref{reference-based}.
We compare them to claimwise annotations, and asses their quality using a Brier score as suggested in~\citealt{rufibach2010use}. 
We measure 0.15 for $Foc$ and 0.12 for $Cov$ (lower Brier score is better), in other research sometimes interpreted~\cite{fernandes} as a \q{superior} agreement.

\subsection{Inter-Annotator Agreement}
The inter-annotation aggreement study is presented in Table~\ref{gold-metrics}.
Due to missing annotations, our first choice for it was the Krippendorff's $\alpha$ coefficient \cite{krippendorff2011computing}.
Its results are, however, abysmally low on our annotations, even reaching negative values reserved for annotation vandalism -- the cause of this are very high levels of class prevalence in our annotations\footnote{Prevalence of the most common class in faithfulness, fluency, atomicity and decontextualisation grading is 0.84, 0.9, 0.98, and 0.91, respectively.}, overly increasing the chance of accidental agreement Krippendorff's alpha targets to punish.
This is a known pitfall of Krippendorff's alpha, also encountered in the~\citealt{wright-etal-2022-generating}'s paper which proposed the reference-free metrics.
To overcome it, we also report the Gwet's Agreement Coefficient 1 (AC1) that was designed to be less sensitive to trait prevalence.~\cite{gwet2010handbook}
We report the values\footnote{Values were computed using Python port of R library \textit{irrCAC} - https://irrcac.readthedocs.io/en/latest/} for AC1 coefficient in table \ref{gold-metrics}.
From the results, we can see that the AC1 coefficient is high in all cases\footnote{The Gwet's AC1 coefficient ranges from 0 to 1.} thus, we conclude that our annotation process was reliable. To provide a complete measurement, we also report \%-agreement -- a proportion of labels on which all annotators agree~\cite{Artstein2017}.

\section{Conclusion}\label{sec:conclusion}
Our study has examined the problematic of automated claim extraction in its full scope: from data gathering (publishing \ds{} dataset) through model training (suggesting low- and high-resource options), benchmarking (publishing a Python framework for automated evaluation) to its validation with annotators, suggesting the hardest and most appropriate score for claim extraction being the $F_{fact}$ we introduce in Section~\ref{ffact}, requiring gold claims for reference.

Interestingly, we have shown that the task of claim extraction can be to a significant extent solved using single-sentence abstractive summarization methods and a diverse beam search decoding strategy, which is particularly interesting for low-resource languages and environments.

Overall, we have aimed this aticle as a breadth-first exploration of the claim extraction task using abstractive methods, pointing directions which to explore in more depth.
\subsection{Future works}

A central challenge to claim extraction is the notion of claim \textit{check-worthiness}, now explored as a classification task, for example by~\citealt{10.1007/978-3-031-28241-6_59}.
Our paper models a dummy version of it, training the models to only focus on the middle sentence claims like FEVER annotators did.
Adapting existing data such as CLEF-CheckThat! datasets for text generation, rather than classification task, could capitalize on our findings to train generative models which only focus on check-worthy claims.

We also encourage the use of our evaluation metric framework to yield additional insights into any system for knowledge-intensive tasks, especially where multiple gold answers can be found or synthesized, such as Summarization or Retrieval-Augmented Generation (RAG).

\section*{Limitations}
The main limitation of the work is that it does not explore a specific sound notion of check-worthiness for measuring the claim \textit{focus} and \textit{coverage}, but model it after the FEVER WF1a annotation objective, which instructs the annotators to extract 2-5 relevant claims from the \textit{middle} sentencence of shown context.
This is a sort of a dummy objective to optimize for (albeit, has its sense, since every sentence in encyclopedic style is supposed to contain some check-worthy claim), and the fact that our SotA model scores $F_{fact} = 0.64$ expressing the extent to which it extracts the same claims human would only motivates the research on \textit{what other objectives of claim relevancy} could the models be trained for, rather than marking the claim extraction challenge as resolved.

So, while we have the approaches and metrics that have been validated in vitro against human gradings of the same task, the generalization of the scheme to claim extraction in the wild remains largely to be explored.

\section*{Ethics Statement}
While our data and evaluation framework seems to be disambiguous enough to use, our proposed method of generating the factual claims using a one-to-many abstractive text-to-text approach raises critical requirements on the deployment of such models.
The models are suitable to produce claims at a scale, and the evaluation framework can pinpoint most of their hallucinations.
However, the use of such models to intepret individual person's claims in language forms such as Tweets and debate transcripts, like~\citealt{barroncedeno2020overview} do in their check-worthy claim detection tasks, is ethically problematic, as the abstractive models have no hard guarantee of aligning to the facts and semantical nuances of the speaker.
The models should therefore be avoided to use without a human in the loop in such scenario, and their use should be disclosed.

Furthermore, we declare that we have used LLM-based code assistants when producing the code supplement of this paper, to generate boilerplate Python code.

\bibliography{custom, anthology}

\clearpage
\appendix

\section{Annotation Platform}
\label{sec:annotation}

\begin{figure*}[h]
  \centering
  \includegraphics[width=\textwidth]{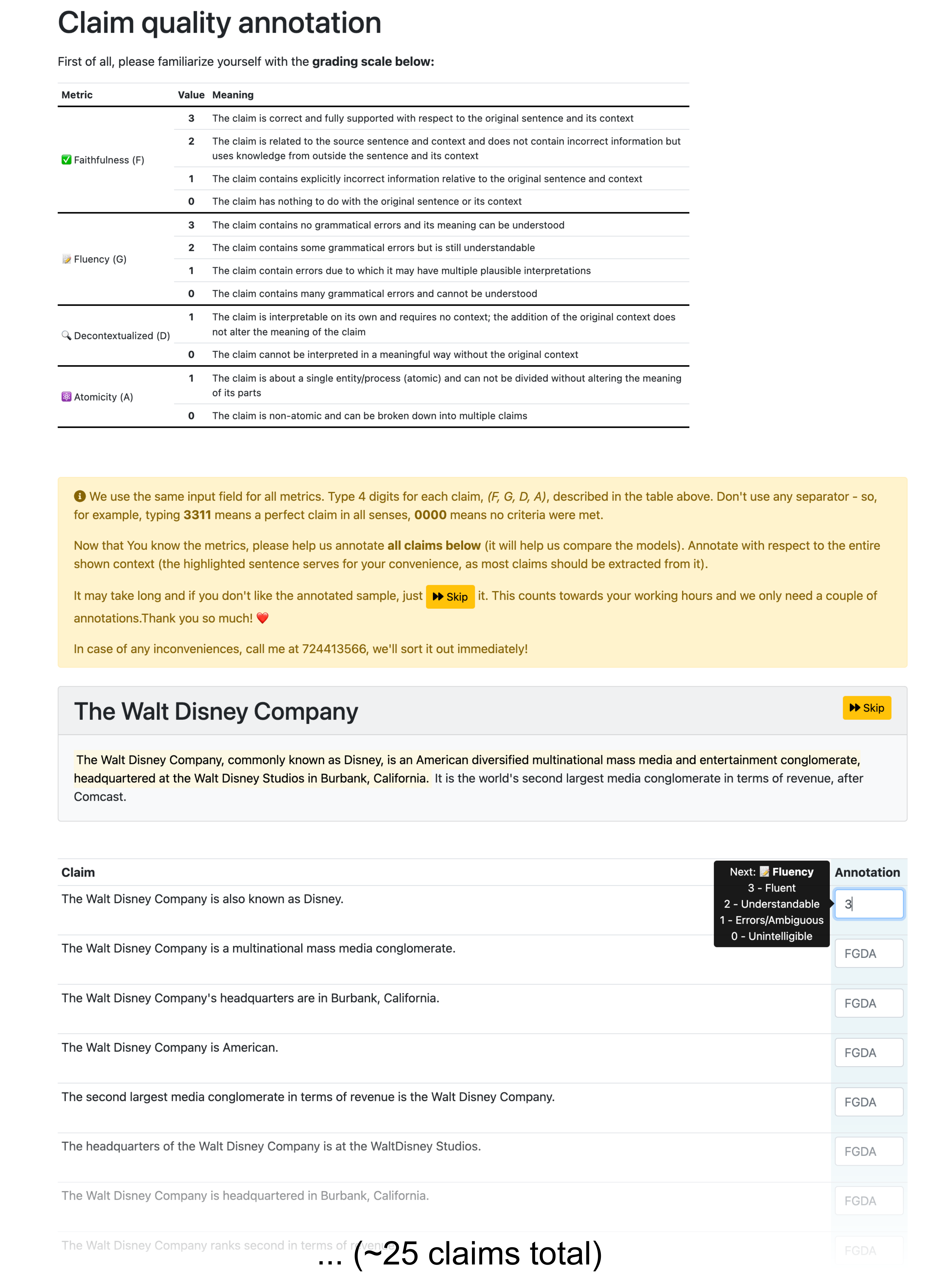}
  \caption[Annotation platform - claim quality annotation screenshot]{Example of the claim quality annotation page. The annotators were asked to grade the quality of the claimss.}
  \label{fig:platform_claim_quality}
\end{figure*}

\begin{figure*}[h]
  \includegraphics[width=\textwidth]{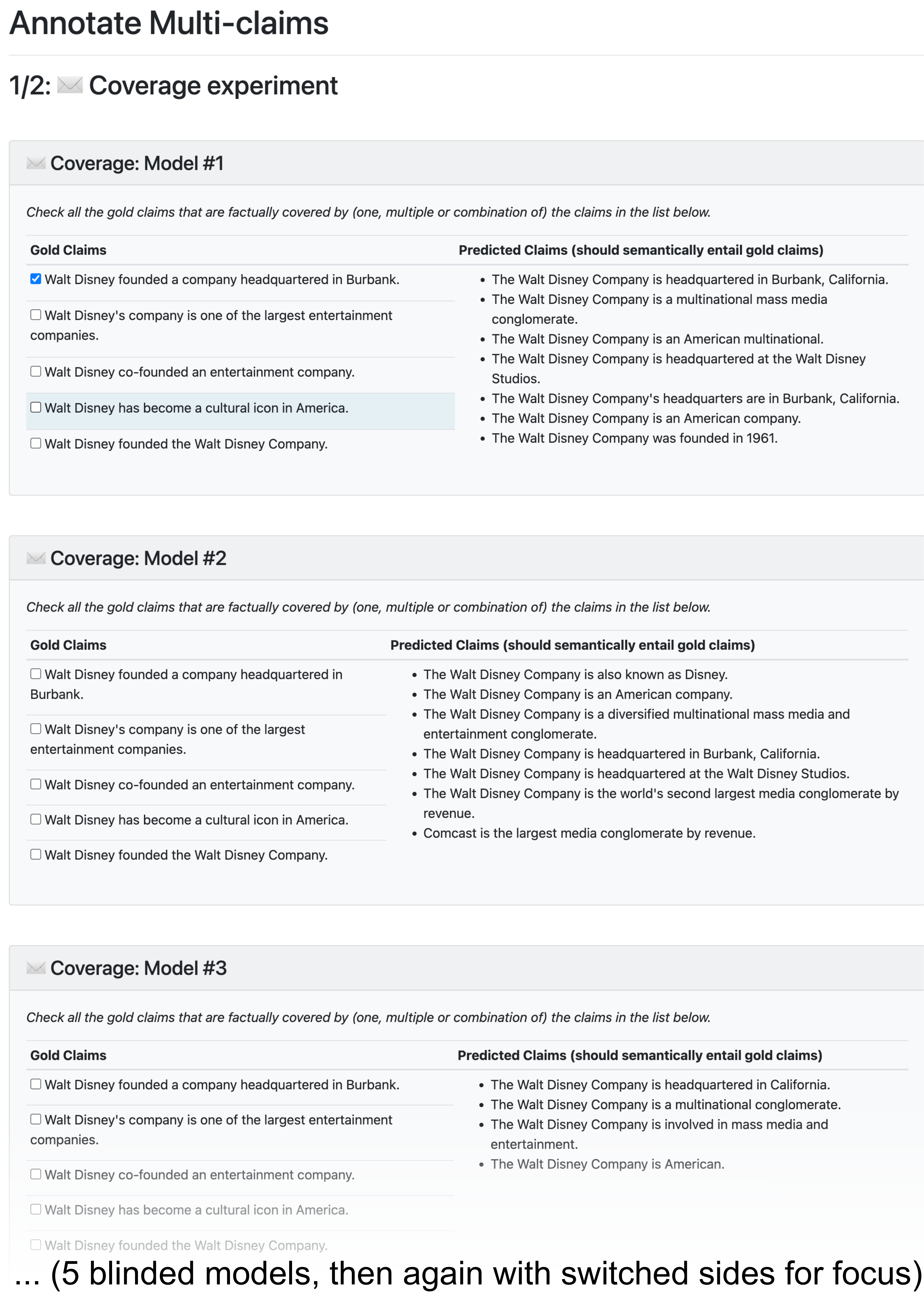}
  \caption[Annotation platform - multi-claim annotation screenshot]{Example of the multi-claim annotation page. The annotators were asked to evaluate the coverage and focus of the claims.}
  \label{fig:platform_multi-claims}
\end{figure*}

To facilitate the human annotations, we have created a PHP annotation platform. It supports the annotation of the quality of the claim (example in Figure~\ref{fig:platform_claim_quality}), multi-claim annotation (example in Figure~\ref{fig:platform_multi-claims}) and also cross-annotations providing data for the inter-annotator agreement evaluation.

\begin{figure}[h]
  \centering
  \includegraphics[width=\columnwidth]{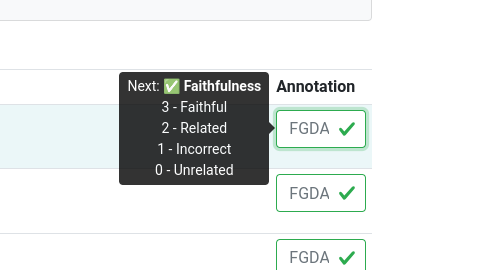}

  \caption[Platform - popup help example]{Example of popup help for the claim quality annotation. This example shows the help for the faithfullness metric and simillar popups were provided also for the other ones.}
  \label{fig:platform_popup}
\end{figure}

The claim quality annotation page first presents the grading scale to let the annotators familiarize themselves with it. Then, a sample was provided with a highlighted sentence from which most of the claims should have been extracted. The claims obtained with various models are presented below. These claims were shuffled to avoid bias towards any model. Moreover, the annotators did not know which model generated/extracted each claim. Given the context sample and a claim, the annotators were asked to grade the quality by assigning scores\footnote{Scores are from range 0-3 for faithfulness and fluency, and either 0 or 1 for contextualized and atomicity.}. To help the annotators fill scores, we have provided a help popup with information on which metric they are considering now. An example of the popups is shown in Figure~\ref{fig:platform_popup}.

The second type of annotation was the multi-claim annotation, where the annotators were first asked to evaluate coverage and then the focus. These annotations were done per model. However, the model type was hidden so as not to induce any bias.

\end{document}